\documentclass[10pt,twocolumn,letterpaper]{article}

\usepackage[accsupp]{axessibility}  
\usepackage{iccv}
\usepackage{times}
\usepackage{epsfig}
\usepackage{graphicx}
\usepackage{amsmath}
\usepackage{amssymb}

\usepackage{xcolor}
\usepackage{adjustbox}
\usepackage{floatrow}
\newfloatcommand{capbtabbox}{table}[][\FBwidth]
\usepackage{array}
\usepackage{multirow}
\usepackage{caption}

\newcommand{\ourmodel}{\textsc{3DStyleNet}}

\usepackage[breaklinks=true,bookmarks=false]{hyperref}

\iccvfinalcopy 

\def\iccvPaperID{6694} 

\ificcvfinal\fi

\begin{document}

\title{3DStyleNet: Creating 3D Shapes with Geometric and Texture Style Variations}

\author{Kangxue Yin$^{1}$ \quad Jun Gao$^{1,2,3}$  \quad Maria Shugrina$^{1}$  \quad Sameh Khamis$^{1}$ \quad Sanja Fidler$^{1,2,3}$ \\
	\quad \small{NVIDIA\textsuperscript{1} \quad University of Toronto\textsuperscript{2} \quad Vector Institute\textsuperscript{3}} \vspace{3pt}\\
	\quad \quad \texttt{\scriptsize \{kangxuey, jung, mshugrina, skhamis, sfidler\}@nvidia.com}
}

\twocolumn[{%
\renewcommand\twocolumn[1][]{#1}%
\maketitle
\begin{center}
    \centering
    \captionsetup{type=figure}
    \vspace{-3mm}
    \includegraphics[width=0.995\linewidth]{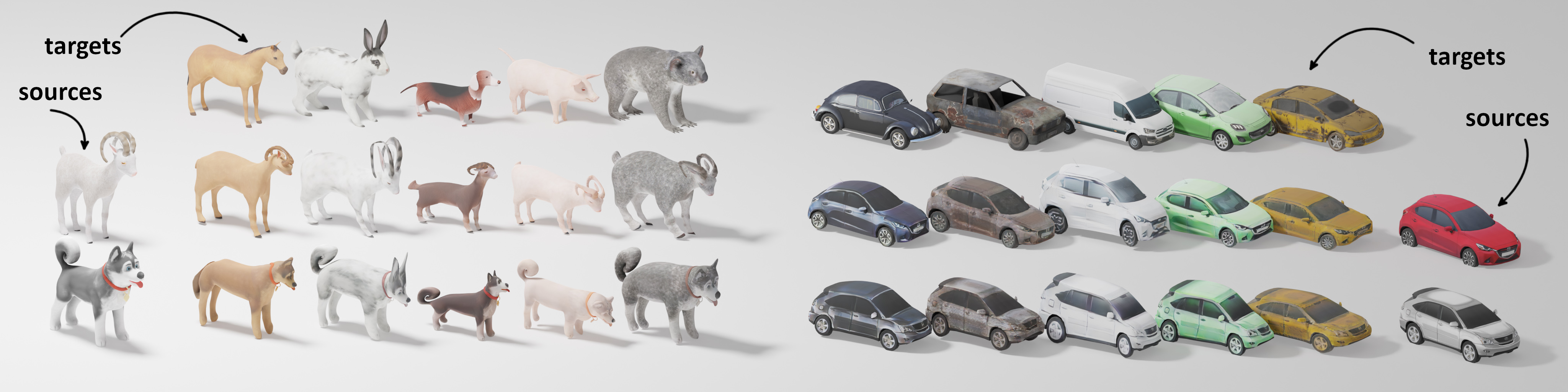}
    \vspace{-2mm}
    \captionof{figure}{\small We propose {\ourmodel}, a neural stylization method for 3D textured shapes. Our method creates novel geometric and texture variations of 3D objects by transferring the shape and texture style from one 3D object (target) to another (source). 
     }
\end{center}%
}]

\ificcvfinal\thispagestyle{empty}\fi

\begin{abstract}
	
We propose a method to create plausible geometric and texture style variations of 3D objects in the quest to democratize 3D content creation.  
Given a pair of textured source and target objects, our method predicts a part-aware affine transformation field that naturally warps the source shape to imitate the overall geometric style of the target. In addition, the texture style of the target is transferred to the warped source object with the help of a multi-view differentiable renderer. 
Our model, {\ourmodel}, is composed of two sub-networks trained in two stages.  
First, the geometric style network is  trained on a large set of untextured 3D shapes. 
Second,  we jointly optimize our geometric style network and a pre-trained image style transfer network with losses defined over both the geometry and the rendering of the result.   
Given a small set of high-quality textured objects, our method can create many novel stylized shapes, resulting in effortless 3D content creation and style-ware data augmentation.
We showcase our approach  qualitatively on 3D content stylization, and provide user studies to validate the quality of our results. In addition,  
our method can serve as a valuable tool to create 3D data augmentations for computer vision tasks. Extensive quantitative analysis shows that {\ourmodel} outperforms alternative data augmentation techniques for the downstream task of single-image 3D reconstruction. 
Project page: \url{https://nv-tlabs.github.io/3DStyleNet/}

\end{abstract}

\vspace{-2mm}
\section{Introduction}
\vspace{-2mm}

The remarkable success of neural image style transfer has demonstrated deep learning as a powerful tool to create artistic images~\cite{gatys2016image,Huang2017ArbitraryST,Sheng2018AvatarNetMZ,Li2017UniversalST,Li2018ACS,Gatys2017ControllingPF,Li_2019_CVPR}, with both casual and professional applications~\cite{stewart17}. 
Although editing 3D content is arguably a more arduous and time-consuming task, which makes automatic tools especially attractive, equally successful formulation of style transfer for the 3D domain has not been proposed. At the same time, the demand for 3D content is growing due to the popularity of gaming, AR/VR, 3D animated films, and simulation of virtual worlds. In our work, we seek a style transfer formulation applicable to \textit{3D content} creation, including both the geometric shape of objects and their color texture. 

Prior works in 3D style transfer address either shape or color stylization, but do not extend to both of these critical attributes of 3D objects. Classical methods consider deformation transfer from one object to another with the guidance of shape correspondence~\cite{sumner2004deformation} and the  transfer of the texture map from one shape to another by minimizing distortion~\cite{schmidt2019distortion,panozzo2013weighted}.
Deep learning methods likewise addressed either geometry~\cite{liu_2019,liu2018paparazzi,Yifan:NeuralCage:2020} or texture stylization~\cite{mordvintsev2018differentiable}. For example,~\cite{liu_2019} proposes an energy optimization framework based on surface normals for geometry cubification, while~\cite{Yifan:NeuralCage:2020} deforms a shape by utilizing a neural network to predict a cage defining a smooth warp of the shape. 
In~\cite{mordvintsev2018differentiable} and~\cite{kato2018renderer}, the style of an artistic painting is transferred to the texture or surface of a 3D object.
None of the above methods is able to perceive and transfer the geometric and texture style jointly from one 3D object to another.

In this paper, we aim to create novel geometric and texture variations of 3D objects by transferring the geometric and texture style from one 3D object to another. Unlike existing approaches, our method performs joint optimization over shape and texture to ensure consistency of the final 3D object.
Our method, referred to as {\ourmodel}, treats simple geometric relationships (e.g., relative scales, positions, rotations) between the semantic parts of a 3D object as a global shape style, which can be abstracted with a set of ellipsoids. We model geometric style transfer with a part-aware affine transformation field, defined based on the ellipsoids, that warps the semantic parts of source shape to be in similar relationships to these parts in target shape.
We design a neural network to perform this task, which we train on a large dataset of 3D shapes. In order to achieve high-quality texture style transfer, a proper alignment of the object geometry is required.  We therefore couple and jointly optimize our geometric style network with a pre-trained image style transfer network~\cite{Li_2019_CVPR} for joint geometric and texture style transfer using losses defined over multi-view rendering produced by a differentiable renderer~\cite{Laine2020diffrast}.

Our {\ourmodel} creates variations of shapes in their style space, yielding a shape creation tool which can be used by naive users for 3D content creation. 
To validate the quality of our results, we conduct a user study which shows that our method can produce higher-quality results than a strong baseline that combines a SOTA shape deformation method and a SOTA image style transfer method.
Furthermore,  we show that our {\ourmodel} can also serve as a 3D data augmentation method for improving the performance of downstream computer vision models. We showcase our approach as a data augmentation strategy for the task of single image 3D reconstruction, demonstrating boosts in performance over strong baseline augmentation techniques.

\vspace{-2mm}
\section{Related work}
\vspace{-1mm}
\subsection{Image Stylization}
\vspace{-1mm}
Artistic stylization is a long-standing research area with a large body of work for images and videos~\cite{Kyprianidis2013StateOT,gooch2001non}. Traditional approaches rely on stroke-based approximations~\cite{hertzmann1998painterly}, region-based motion transfer~\cite{agarwala2004keyframe}, and hand-crafted local image features~\cite{Hertzmann2001ImageA}.
The classic \emph{Image Analogies} method of Hertzmann \etal~\cite{Hertzmann2001ImageA} laid the groundwork for  Neural Style Transfer pioneered by Gatys et al.~\cite{gatys2016image}. This has become a popular research direction in recent years~\cite{jing2017neural}, with the majority of Neural Style Transfer work targeting images and videos. We build on methods in the image domain~\cite{Li_2019_CVPR} to target 
style transfer for textured 3D 
shapes. 

\vspace{-1mm}
\subsection{Shape Editing and Stylization}\label{ssec:related:geomtransfer}
\vspace{-1mm}
\textbf{Shape Deformation and Modeling.} A broad class of techniques allow manipulation of 3D surface geometry,
for example, using controllable handles and skeletons~\cite{sorkine2007rigid,sumner2007embedded,gal2009iwires},
governed by closed-form solutions such as linear blend skinning (LBS)~\cite{lindholm2001user} or energy minimization techniques~\cite{sorkine2007rigid,sumner2007embedded}. Unlike
these interactive methods, deformation transfer seeks to
automatically transfer a series of poses of a source shape to the target shape,
using provided correspondences and optimization~\cite{sumner2004deformation} or by fitting
a neural network~\cite{Yifan:NeuralCage:2020,gao2018automatic}. The goal
of our work is not to transfer deformations, but rather 
to stylize the input shape via a part-aware affine transformation on geometry. Compared to the recent method
Neural Cage~\cite{Yifan:NeuralCage:2020} (which stylizes geometry only), we achieve superior performance experimentally.

\textbf{Geometric Stylization.} Several approaches seek
to stylize geometric features of a shape. Liu and Jacobson~\cite{liu_2019}
adapt the As-Rigid-As-Possible energy~\cite{sorkine2007rigid} with $L_1$ regularization on surface normals to warp an input mesh 
 into the style of a cube while maintaining shape details. Others propose
 automatic stylization in the style of 3D collages~\cite{gal20073d,theobalt2007animation}, \textit{LEGO}~\cite{luo2015legolization}, furniture~\cite{xu_siga10,Hu2017Style} or Manga~\cite{shen2012sd}. Style has also been considered
 on a more local level. Mesh smoothing or filtering, using  classical operators~\cite{sorkine2005laplacian} or neural networks~\cite{takikawa2021neural}, can also be viewed as a style where one progressively changes the level-of-detail while maintaining geometry-aware features~\cite{hoppe1996progressive,stein2018developability}. Recent works also propose to learn local 3D textures~\cite{hertz2020deep} or train style-specific mesh
 refinements~\cite{liu2020neural}. Our approach learns deformations that stylize the global object shape in a plausible way given a target shape as a guidance.

\textbf{Image-Based Approaches.} Various recent works follow a  2D-to-3D approach, leveraging a pretrained style network~\cite{gatys2016image} and differentiable rendering to minimize the style energy in the rendered image space~\cite{liu2018paparazzi,kato2018renderer}. Parparazzi~\cite{liu2018paparazzi} focuses on editing the local geometric details, while ignoring the style of global deformations. N3MR~\cite{kato2018renderer} edits both texture and geometry towards the style of a target 2D image. Our aim in this work is to transfer the style  from one 3D object to another. 

\begin{figure*}[t!]
	\includegraphics[width=0.9\linewidth]{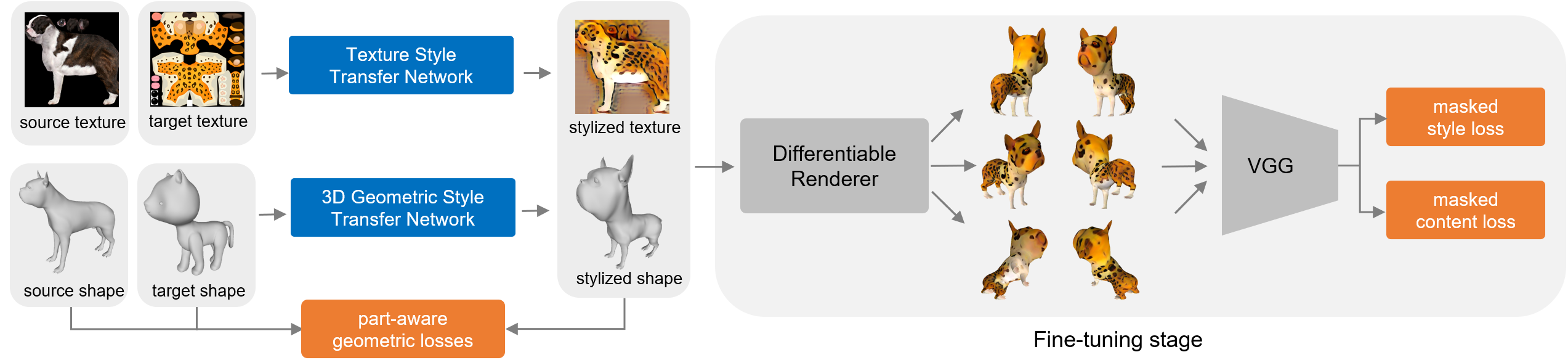}
	\vspace{-3mm}
	\caption{\small An overview of {\ourmodel}, comprised of two main modules: a geometric style transfer network and a texture style transfer network. Each module is pre-trained on either 3D shape transfer or image stylization. We then perform joint geometry and texture optimization by utilizing a differentiable renderer. 
		\vspace{-5mm}}
	\label{fig:overview}
\end{figure*}

\vspace{-1mm}
\subsection{Texture Transfer}\label{ssec:related:textransfer}
\vspace{-1mm}

The problem of
transferring texture from one 3D shape to another 
can be solved by finding a dense mapping from one
geometry to another. A number of geometry processing
methods for computing such maps have been proposed \cite{schmidt2020inter,edelstein2020enigma}, but often require
user-provided correspondences \cite{ezuz2019elastic}  or place strong assumptions
on the input geometry. For example, most meshes found
in the wild are not manifold \cite{edelstein2020enigma}, watertight \cite{schmidt2020inter} or homeomorphic to a disk \cite{schmidt2019distortion}.
More importantly, these purely geometric approaches lack
awareness of semantic features, like eyes of a character, or 
the ability to modify the texture itself in order to
preserve geometric patterns when applied
to a vastly different geometry.
In contrast, our method employs a differentiable
renderer and content-aware losses that allow 
robust texture transfer
that respects both the content and the higher frequency
patterns of the texture. Our method is applicable
to any triangle mesh including non-manifold meshes with
holes and multiple objects. For a related task of image to 3D model texture transfer,
our method can be used to directly replace the need for 
a 3D proxy or lighting-texture decomposition as proposed in \cite{wang2016unsupervised}.

\vspace{-2mm}
\section{Method}
\vspace{-2mm}
The goal of our method is to synthesize variations of 3D objects in their geometric and texture style space. 
Our method takes as input a \emph{source} 3D model $P$ representing the ``content", and a \emph{target} 3D model $Q$ representing the desired ``style". Unlike Neural Style Transfer for images, we \emph{edit} the source shape rather than generating new content, thus preserving the level of detail of the source.
In addition to surface geometry, our method requires a texture map of the target and, if available, of the source. Unlike prior works, {\ourmodel} performs both geometric and texture stylization to achieve a wide range of 3D variations. 

There is no single way to define the style of a 3D object. In our work, we consider global geometric style on the semantic part-level. For example, a cartoon dog may have a much larger head than its realistic counterpart. 
Our stylization leaves local geometric details of the source unchanged, which is often desirable for high-quality models. Instead, 
we treat geometric relationships between semantic parts, such as relative scales and positions, as the object's style. We abstract this style with a set of ellipsoids and model the geometric style transfer with a learned 3D part-aware affine transformation field that warps the semantic parts.
We define texture style of a 3D object similarly  to modern image style transfer techniques,
and perform stylization in the rendered image space by leveraging differentiable rendering. 

The {\ourmodel} is comprised of two main network components,  the Geometric and Texture Style Transfer Networks, as shown in Figure~\ref{fig:overview}. We first pre-train the 3D  Geometric Style Transfer Network (\S\ref{ssec:method:geostyle}) on a set of untextured shapes and the Texture Style Transfer Network (\S\ref{ssec:method:texture}) on image datasets. To make them work jointly for geometric and texture style transfer, we perform joint geometry and texture optimization (\S\ref{ssec:method:finetune}). A large set of 3D object variations can be created automatically by applying the joint optimization for all object pairs in a set of 3D objects, or in a
user-driven design tool.

\vspace{-2mm}
\subsection{3D Geometric Style Transfer Network }\label{ssec:method:geostyle}
\vspace{-2mm}

\begin{figure}[t!]
	\includegraphics[width=1\linewidth]{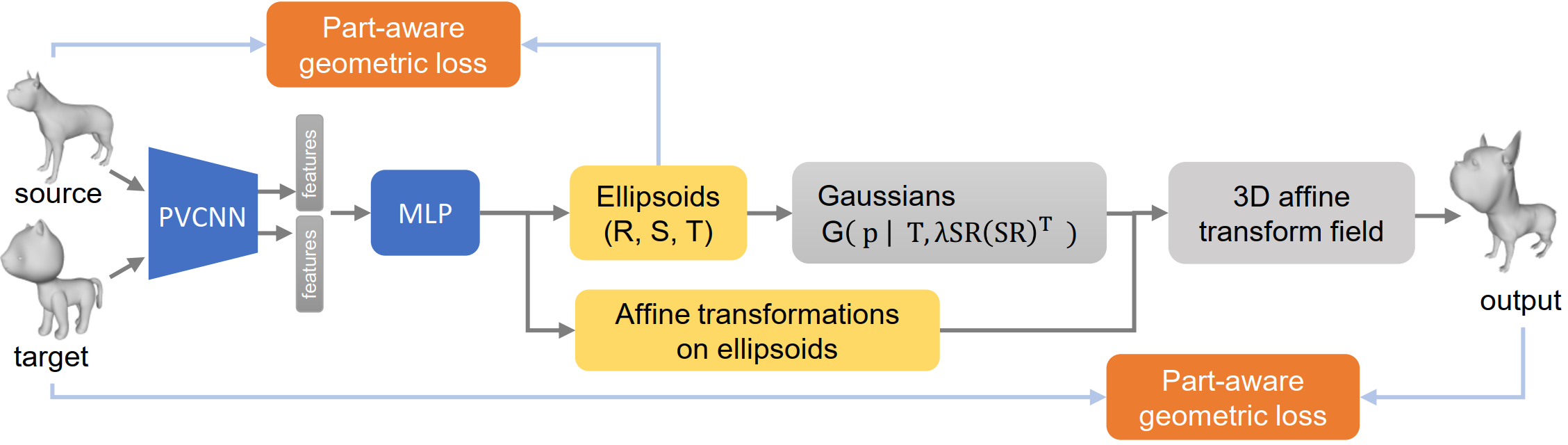}
	\caption{\small Our part-aware 3D Geometric Style Transfer Network.\vspace{-5mm}}
	\label{fig:geoNet}
	\vspace{-3mm}
\end{figure}

We define the geometric style of a 3D shape by focusing on relationships between semantic parts. For example, an adult tiger has a relatively smaller head and longer legs than a baby tiger. Such global geometric style is well represented by approximate shapes of semantic parts. To modify geometric style given target guidance, our 3D Geometric Style Transfer Network (Figure~\ref{fig:geoNet}) learns to predict a part-aware affine transformation field regressed from a source and target shape pair. This predicted
affine transformation field can then be used to smoothly deform the
vertices of the source shape while preserving fine geometric details.

\begin{figure}[t!]
\centering
\includegraphics[width=\textwidth]{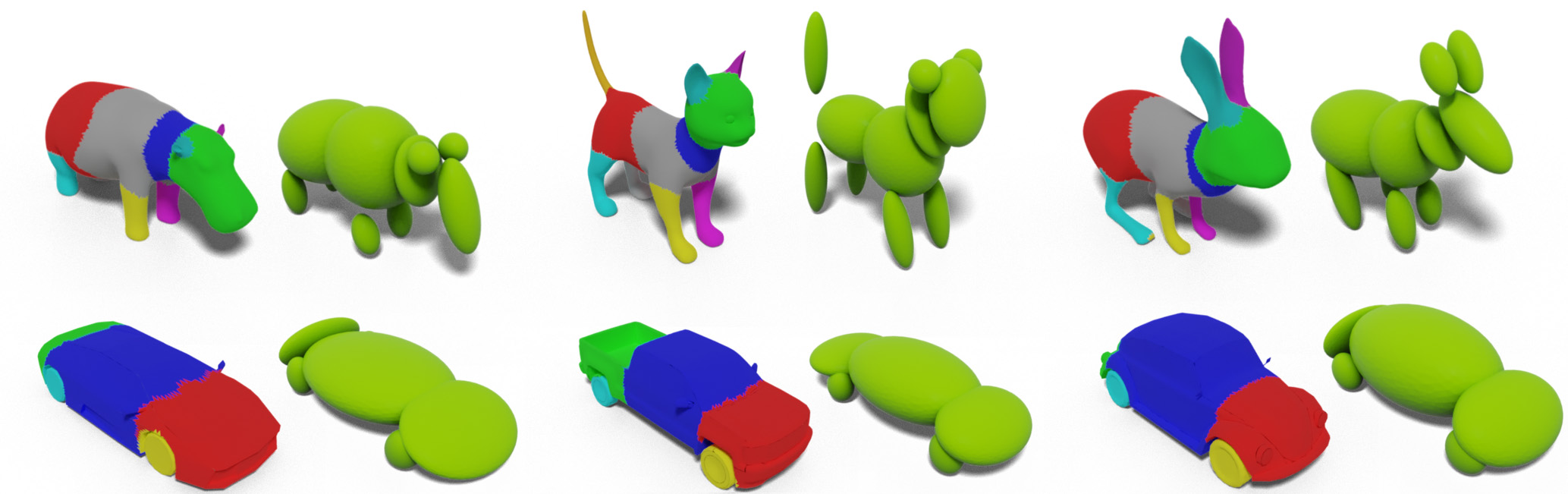}
\vspace{-5mm}
\caption{\small Example part segmentation and predicted ellipsoids for the same shapes.\vspace{-5mm}}
\label{fig:semi_part_and_affine}
\vspace{-2mm}
\end{figure}

\textbf{Network Architecture:} Our geometric network 
takes point cloud samples of the two shapes as input. We use PVCNN~\cite{liu2019pvcnn} to encode the point clouds, and feed the concatenated embeddings to an MLP with skip connections. The network is defined for a fixed number of semantic parts $N$. For each semantic part $i$ in the source shape, the MLP outputs parameters of an ellipsoid $E_i$ that best approximates the part (see Figure~\ref{fig:semi_part_and_affine}),
and a 3D affine  transformation $A_i$ 
for warping the part. This output is then used to compute a smooth affine transformation field to deform the source geometry, while preserving local details.

\textbf{Part-Aware Transformations:} 
We use the parameters of the predicted ellipsoids $E_{1},\dots,E_{N}$ to compute smooth skinning weights for any points on the source shape. 
Using these skinning weights and the $N$ predicted affine transformations,
we compute the deformation of any source point using the LBS model~\cite{lindholm2001user}. Abstractly, the predicted
ellipsoids represent ``what" to deform, and the affine transformations represent ``how" to deform it. 
To derive the skinning weights, we
observe that an ellipsoid $E_i$ can  also be represented by a 3D affine transformation 
composed of $S_i$, $R_i$ and $T_i$ that scales, rotates and translates a unit sphere to turn it into an ellipsoid. With this decomposed representation of the ellipsoid $E_i$, we can define a 3D Gaussian that aligns with it:
\vspace{-1mm}
\begin{equation}
g_i(p) = G(p | T_i, \lambda S_iR_i(S_iR_i)^{T} )
\end{equation}
where $p$ is a 3D point, 
$T_i$ is the mean of the Gaussian and center of the ellipsoid, $\lambda S_iR_i(S_iR_i)^{T_i}$ is the covariance matrix, and $\lambda$ is a fixed scalar for controlling the spread (we use $\lambda=4.0$). 
The $N$ functions $g_i$ derived from $N$ part ellipsoids define an $N$-channel 3D blending field. We use this normalized blending field to interpolate the affine transformations $A_i$ of all semantic parts to obtain a single 3D affine transformation field $\phi$.  We warp the source shape $P$ with the affine transformation field to obtain the stylized output shape $\phi(P)$. In practice, Gaussians and the blending field are only evaluated for every vertex of the source mesh.

\textbf{Part-Aware Losses:} 
We train the 3D Geometry Style Transfer Network using part-aware geometric losses, requiring semantic part labeling of the input dataset. To this end,  we manually labeled a small set of training shapes, and train a BAE-NET~\cite{chen2019bae_net} under hybrid supervision
to predict the semantic part labels for all shapes.  
See Supplement. for details in training the segmentation network. 
In Fig~\ref{fig:semi_part_and_affine}, we visualize segmentation results on a few samples. In total, we annotated $N=11$ semantic parts for animals,  $7$ parts for cars, and $6$ parts for people. 
Overall, we observed the predicted segmentation quality to be high, but smaller parts like ears were sometimes noisy. 

The main component of our loss function is the part-aware distance between shapes $P$ and $Q$, defined as:
\vspace{-1mm}
\begin{equation}
D_{part}(P, Q) =  Ch_{L1}(P, Q) +  \sum_{i \in L}Ch_{L_1}(P_i, Q_i)
\end{equation}
where $P, Q$ are the sampled point sets and $P_i$ and $Q_i$ denote the point subsets for part $i$,
and $Ch_{L1}(P, Q)$ is the averaged $L_1$-Chamfer distance. Including both global (first term) and part-wise (second term) Chamfer distances makes this loss more robust to segmentation noise.

The final loss function for our Geometric Style Transfer Network is:
\vspace{-1mm}
\begin{equation}\label{eq:geoloss}
\begin{split}
Loss(P,Q, \phi) & = D_{part}(\phi{(P)}, Q)   + D_{part}(\xi(P), P)  \\
               & +  \alpha (D_{sym}(\phi{(P)}) + D_{sym}(\xi{(P)}) )
\end{split}
\end{equation}
where $\xi(P)$ is the sampled surface points of all ellipsoids $\{E_i\}$ predicted for source shape $P$,  and $\phi{(P)}$  is the warp of the source shape as described above. $D_{sym}(\phi{(P)})$ is an optional symmetry regularization term defined as the averaged $L1$-Chamfer distance between $\phi{(P)}$ and its reflection over its plane of symmetry (we use $\alpha=0.1$).

\textbf{Training:} Due to the limited availability of high-quality textured 3D shapes, we designed the Geometric Style Transfer Network to be trained on a large set of shapes without textures.  Similarly to~\cite{Yifan:NeuralCage:2020}, we run self-supervised training by enumerating all possible pairwise combinations of the training shapes. We use the Adam  optimizer for 40,000 iterations, with a batch size of 32. The initial learning rate is 0.001, and is halved after every 20\% iterations. 

\vspace{-2mm}
\subsection{Texture Style Transfer Network}\label{ssec:method:texture}
\vspace{-2mm}

To transfer texture style, we reuse a linear image style transfer network~\cite{Li_2019_CVPR}, with texture images (Figure~\ref{fig:overview}) as the source
and target. The innovation of {\ourmodel} is
applying this component to 3D models using a differentiable renderer in the fine-tuning stage (\S\ref{ssec:method:finetune}). This allows the Texture Style Transfer 
Network to gain awareness of geometric properties, absent in unlabeled texture images alone.

\textbf{Training:} This network is pre-trained on images. Specifically, the encoder is part of the VGG-19~\cite{simonyan2014very} trained on ImageNet~\cite{deng2009imagenet}, and the decoder is trained on MS-COCO~\cite{coco}. The linear transformation module is trained on MS-COCO and WikiArt~\cite{wikiart} as the content and style image sets, respectively. Please refer to ~\cite{Li_2019_CVPR}  for details. 

\vspace{-2mm}
\subsection{Joint Geometry and Texture Optimization}\label{ssec:method:finetune}
\vspace{-2mm}

Due to the lack of textured training shapes, the Geometric and Texture Style Transfer networks are trained separately on different datasets. 
As the texture network is not 3D-aware, it is not able to avoid seams at the 3D surface regions where uv mapping is discontinuous, and is not able to learn to ignore the black background  in the texture image (e.g., see input texture images in Fig.~\ref{fig:overview}). To overcome these issues, we render the shapes with textures, and jointly optimize the  geometry and texture networks at test time.

Specifically, at test time, we fine-tune both networks for a specific source and target pair. To accomplish this, we render the textured stylized object in multiple views with a rasterization-based differentiable renderer, Nvdiffrast~\cite{Laine2020diffrast},  and evaluate the masked content and style loss on the multi-view rendering for joint optimization of geometry and texture. The two networks learn to work together to  both hallucinate textures that respect target patterns and colors and source boundaries, and to adjust geometric transformations to make texture transfer easier. 

Concretely, we jointly optimize the parameters of the
MLP in the geometry style network (\S\ref{ssec:method:geostyle}) and 
of the linear transform module of~\cite{Li_2019_CVPR} (\S\ref{ssec:method:texture}) to minimize losses defined over both the geometry and multi-view rendering of the stylized object, using the following objective function: 
\vspace{-1mm}
\begin{equation}\label{eq:finetune}
	\small
	\begin{split}
		f(\phi, \hat{m}_P)  &= Loss(P,Q, \phi)  \\
		& +  \beta \sum_{v}{L_{content} [  F_v(\phi(P), \hat{m}_P), F_v(P, m_P)  ]   } \\
		& +  \gamma \sum_{v}{L_{style}[F_v(\phi(P), \hat{m}_P), F_v(Q, m_Q)] }
	\end{split}
\end{equation}
where $m_P$ and $m_Q$ are the texture images of the source $P$ and target $Q$, and $\hat{m}_P$ is the stylized texture image (the source uv map is kept fixed). $F_v(P, m_P)$ is the set of multi-level VGG features of the rendered pixels for shape $P$ with texture $m_P$ under camera view $v$. Crucially, we use background mask output by the renderer to mask out features computed from
irrelevant background pixels. $Loss$ is from Eq.\ref{eq:geoloss}, and style and content losses on the pixel features are defined in \cite{Li_2019_CVPR}.   
We use $\beta=0.01$ and  $\gamma=0.001$ in our experiments.
More details about the joint optimization is provided in Supplement.

\textbf{Fine-tuning Time:} This fine-tuning is fast and typically converges after about 20 steps for each input pair at test-time. Given meshes with 20K faces and $512\times512$ texture images, it takes roughly 9$\sim$10 seconds to get the result on an Nvidia RTX 2080ti GPU.

\vspace{-2mm}
\section{Experiments}
\label{sec:results}
\vspace{-2mm}

In  this section, we evaluate our {\ourmodel} qualitatively and quantitatively, and compare it to alternative baseline approaches. Throughout the experiments, we use the same default hyper-parameters for our networks. 
We showcase our method in stylizing 3D shapes for the animal, car and people categories. We also evaluate our method as a 3D data augmentation technique. 

\vspace{-2mm}
\subsection{Dataset}\label{ssec:exp:dataset}
\vspace{-2mm}

\textbf{Animals:}  
The animal dataset is collected from TurboSquid\footnote{\url{https://www.turbosquid.com/}}. We collected 1,120 non-textured shapes for training our geometric style transfer network. As described in \S\ref{ssec:method:geostyle}, we selected 32 shapes which we manually labeled with semantic parts. We then train a BAE-NET with hybrid supervision to segment all 1,120 shapes into semantics parts. More details of the segmentation method can be found in the Supplement.
In addition to the 1,120 non-textured shapes, we also collect 442 textured animal shapes from TurboSquid for joint geometry and texture style transfer. We screened the dataset to make sure there are no overlapping examples between the non-textured and the textured sets. We use half of the 442 animal set for generating style variations with our method, and use the remaining 221 objects for quantitative evaluation in the data augmentation experiment in \S\ref{ssec:exp:dataaug}. 

\textbf{Cars:}
We use 898 cars from the ShapeNet part segmentation challenge~\cite{yi2017large,chang2015shapenet} for training our geometric style network. In order to get desired part segmentation,
we manually labeled 8 examples, and train a BAE-NET in a similar way to Animals. 
For joint geometric and texture style transfer, we collected 436 textured cars from  TurboSquid since objects on this website come with much higher quality textures than ShapeNet objects. We use 218 for synthesizing style variations, and reserve the remaining 218 for  
quantitative evaluation for the data augmentation experiment.

\textbf{People:}
We collected 500 textured 3D human models in T-pose from RenderPeople\footnote{\url{https://renderpeople.com/}},
and randomly split them into a training set of size 400 and a test set of size 100. To test our method in transferring cartoon style to real human shapes, 
we further collected 20 textured cartoon character shapes from TurboSquid for testing.  We manually labeled 4  examples in the training set for segmentation with BAE-NET.

\vspace{-1mm}
\subsection{3D Style Transfer Results}
\vspace{-2mm}

\begin{figure*}[t!]
	\centering
  \includegraphics[width=0.8\linewidth]{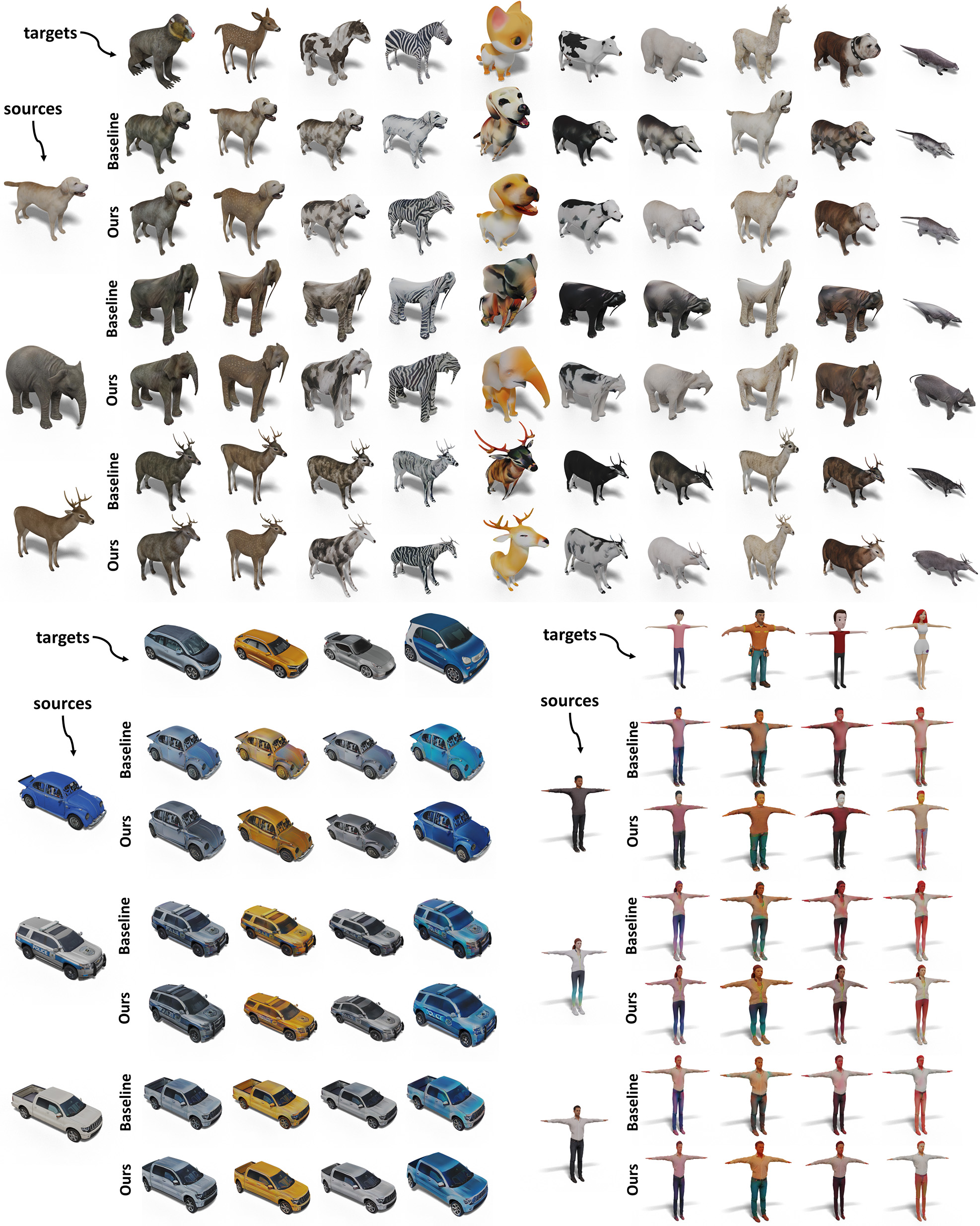}
  \vspace{-0mm}
	\caption{\small \textbf{Qualitative comparison:} Our method v.s. NeuralCage~\cite{Yifan:NeuralCage:2020} + Linear Image Style Transfer~\cite{Li_2019_CVPR}. Notice that our method better captures the style in both geometry and texture. See for example the 5th column of the animal subset. While the baseline simply enlarges the dog's head, our method jointly stylizes both geometry and texture to achieve the cartoon look of the target object.
	\vspace{-5mm} }
	\label{fig:results:qualitative}
	\vspace{-3mm}
\end{figure*}

3D style transfer is one of the main applications of {\ourmodel}. We compare our method to a strong baseline method which combines NeuralCage~\cite{Yifan:NeuralCage:2020} and linear image style transfer network~\cite{Li_2019_CVPR}. Specifically, in the baseline method, we train NeuralCage on our training shapes, and use it to deform the source shape to match the target shape. We then feed the source and target texture images to the pre-trained linear image style transfer network to transfer the texture style as well. 

The results of the baseline and our method are shown in Fig.~\ref{fig:results:qualitative}. Our method can better approximate both the geometric and texture style of the target object. Notice that the baseline largely preserves the source shape for the car category, while ours produces results closer to the target shape. 
Likewise, for animals, our method captures style of the target shape and texture, which is especially evident for the 5th column -- while the baseline simply enlarges the dog's head, our method jointly stylizes both geometry and texture to achieve the cartoon look of the target shape. We also observe some failure cases. For example, the elephant nose in the 6-7th columns is unnaturally warped.
This is because we do not define a nose part in our segmentation, and our method relies on semantic part segmentation of the source object.
The human cartoon style transfer results show that our method does a better job in semantic-ware style transfer than baseline. For example, in the last column, the baseline mistakenly transfers the color of hair to human faces while our method does not.

\textbf{User Study.}
We conducted a user study through Amazon Mechanical Turk where users were asked to compare our results against using Neural Cage~\cite{Yifan:NeuralCage:2020} and Style Transfer~\cite{Li_2019_CVPR} combined. We generated 2000 videos of the source and target models, which we labeled original animals \textit{A} and \textit{B}, and the two stylized models, which we labeled hybrid animals \textit{C} and \textit{D}, where in half the videos our model was hybrid \textit{C} and in the other half it was hybrid \textit{D} to overcome what is known as left-side selection bias.
When asked whether the stylized animal models are closer to the target shape or closer to the source shape, 41.9\% of users reported that our model is closer to the target than it is to the source, compared to only 26.8\% for the baseline model. 69.6\% of users thought our generated colors and patterns are closer to animal B than those of the baseline, and 63.4\% reported that our model has more similar shape and body proportions to the target than those of the baseline. However, only 51.2\% reported that our model is more unique than the baseline and 53.1\% reported that our overall shape is more of a blend of the two shapes compared to the baseline. The questions we asked in the AMT survey can be found in the Supplement.

\begin{figure}[t!]
\vspace{-2mm}
	\centering
	\includegraphics[width=0.97\columnwidth]{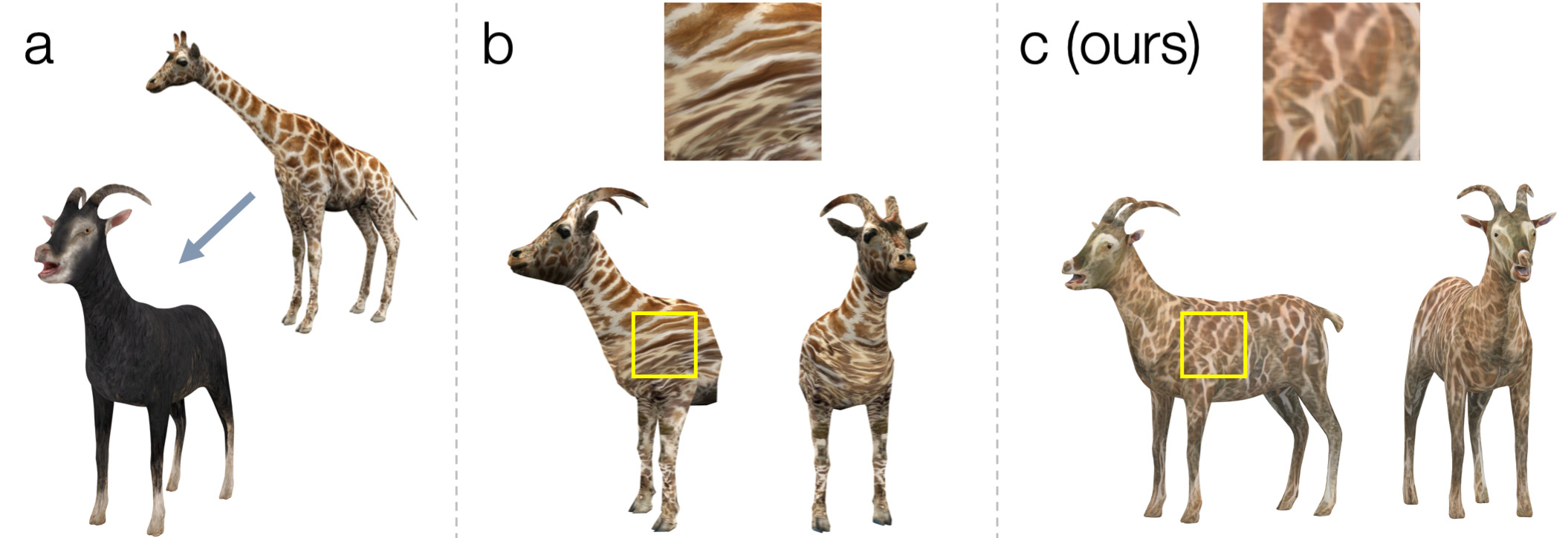}
	\caption{\small \textbf{Texture Transfer:} Transferring
		giraffe texture to a goat model (a), using method of \cite{schmidt2019distortion} (b), and our method (c). 
		\vspace{-5mm}}
	\label{fig:results:tex_transfer}
	\vspace{-3mm}
\end{figure}

\textbf{Mesh Texture Transfer.}
If disabling the geometric style transfer, 
our {\ourmodel} can also be applied to the task of mesh texture transfer (see \S\ref{ssec:related:textransfer}). 
Unlike purely geometric
approaches that seek to find a distortion-minimizing mapping between
geometries, our approach is able to hallucinate new texture images.
As a result, our method avoids
distortions of the source texture caused by widely different target geometry (see
close ups in Fig.~\ref{fig:results:tex_transfer}), and tends to preserve
salient features of the target (such as the face of the goat).
Compared to the baseline, our method is orders of magnitude faster (9$\sim$10 seconds on RTX 2080ti) and robust to non-manifold
non-watertight meshes with diverse topologies, which makes it practical
for processing meshes in the wild. In contrast, competitive methods in \S\ref{ssec:related:textransfer} place many requirements on the input
geometry -- for example, only half of the goat is available for~\cite{schmidt2019distortion} in Fig.\ref{fig:results:tex_transfer} 
because this method is designed for geometries that are homeomorphic to a disk.

\begin{table*}[t!]
    \addtolength{\tabcolsep}{-3.0pt}
    \begin{adjustbox}{width=\linewidth}
        \begin{tabular}{|l||c|c|c||c|c|c||c|c|c|}
            \hline
            \multirow{2}{*}{Augmentation Method} & \multicolumn{3}{c||}{Chamfer $\downarrow$ } & \multicolumn{3}{c||}{Chamfer L1 $\downarrow$ } & \multicolumn{3}{c|}{F-score $\uparrow$}  \\ \cline{2-10}
               & Seen Shapes & Similar Shapes & Unseen shapes & Seen Shapes & Similar Shapes & Unseen shapes &Seen Shapes & Similar Shapes & Unseen shapes \\
              \hline

              No Data Augmentation & 0.025 & 0.035 & 0.065 & 0.073 & 0.102 & 0.187 & 0.323 & 0.210 & 0.085\\
               \hline
               Random Affine &0.040 & 0.043 & 0.053 & 0.114 & 0.123 & 0.152 & 0.211 & 0.195 & 0.144\\
              Neural Cage~\cite{Yifan:NeuralCage:2020} Geometry &0.017 & 0.022 & 0.044 & 0.050 & 0.064 & 0.128 & 0.415 & 0.346 & 0.143\\
               \hline
              Random COCO~\cite{coco} Texure& 0.014 & 0.021 & 0.045 & 0.040 & 0.061 & 0.131 & 0.572 & 0.361 & 0.140\\
              Style Transfer~\cite{Li_2019_CVPR} Texture & 0.015 & 0.023 & 0.047 & 0.042 & 0.068 & 0.136 & 0.518 & 0.316 & 0.122\\
               \hline
              Neural Cage~\cite{Yifan:NeuralCage:2020} + Style Transfer~\cite{Li_2019_CVPR}&0.019 & 0.022 & 0.040 & 0.053 & 0.062 & 0.116 & 0.423 & 0.370 & 0.181\\
              \hline
              \hline
              Ours: Geometry Transfer only& 0.018 & 0.021 & 0.040&0.051 & 0.062 & 0.116&0.423 & 0.358 & 0.188\\
              Ours: Texture Transfer only& \textbf{0.012} & 0.024 & 0.051 & \textbf{0.034} & 0.070 & 0.148 &\textbf{0.623} & 0.314 & 0.118\\
              Ours(w/o finetune): Texture  + Geometry Transfer&0.019 & 0.022 & 0.039&0.054 & 0.063 & 0.111& 0.414 & 0.360 & 0.183\\
              Ours: Texture + Geometry Transfer& 0.016 & \textbf{0.019} & \textbf{0.037}&0.047 & \textbf{0.055} & \textbf{0.107}&0.449 & \textbf{0.394} & \textbf{0.218}\\
            \hline
        \end{tabular}
         \end{adjustbox}
         \vspace{-3.5mm}
        \caption{\small Quantitative results on the downstream task of {\bf Single Image 3D reconstruction} using DISN~\cite{disn} as the 3D reconstruction method and {\ourmodel} compared with baselines as a 3D data augmentation strategy. }
        \label{tbl:single-image-3d-chamfer}   
\end{table*}

\begin{figure*}[t!]
\vspace{-3mm}
\centering
\includegraphics[width=0.1\textwidth,trim=0 50 0 50,clip]{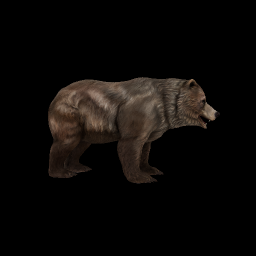}
\includegraphics[width=0.89\textwidth,trim=0 100 0 100,clip]{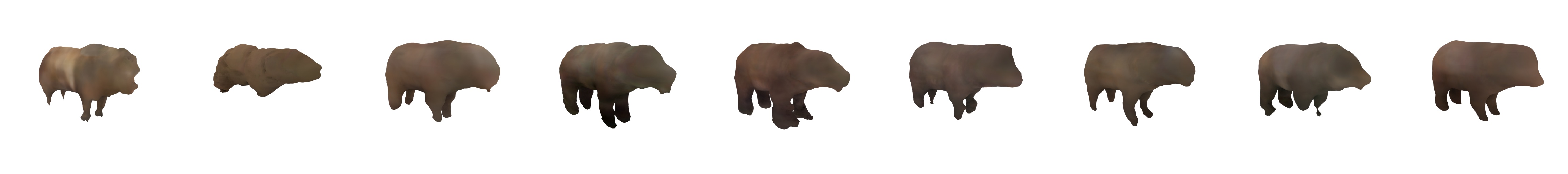}
\includegraphics[width=0.1\textwidth,trim=0 50 0 50,clip]{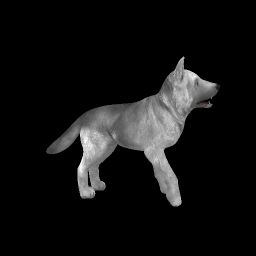}
\includegraphics[width=0.89\textwidth,trim=0 100 0 100,clip]{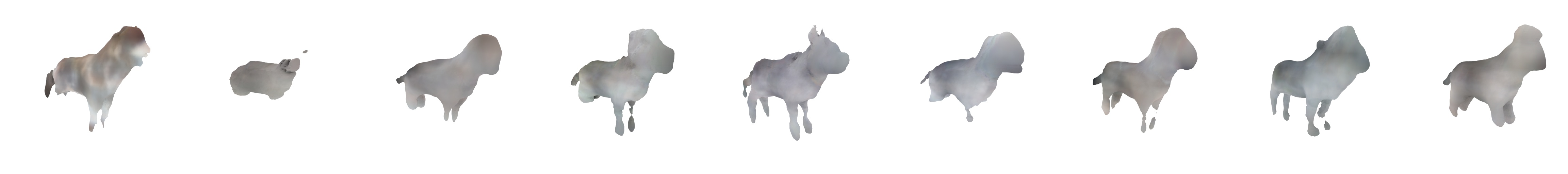}
\includegraphics[width=0.1\textwidth,trim=0 50 0 50,clip]{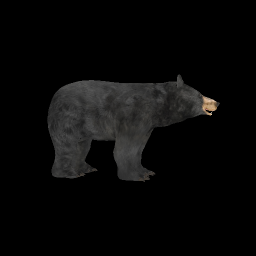}
\includegraphics[width=0.89\textwidth,trim=0 100 0 100,clip]{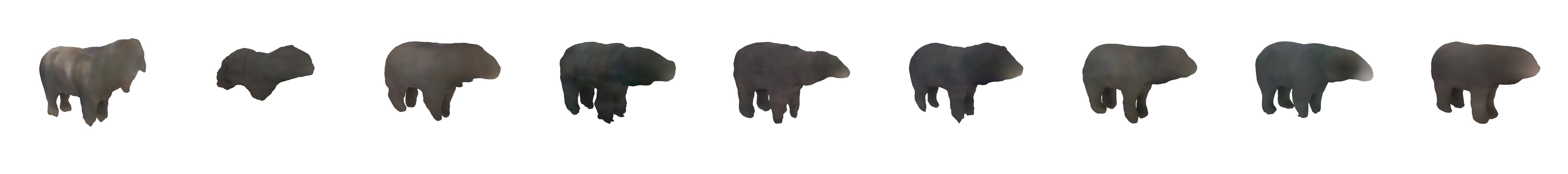}
\includegraphics[width=0.1\textwidth,trim=0 50 0 50,,clip]{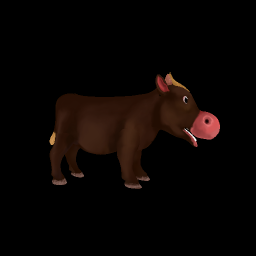}
\includegraphics[width=0.89\textwidth,trim=0 100 0 100,clip]{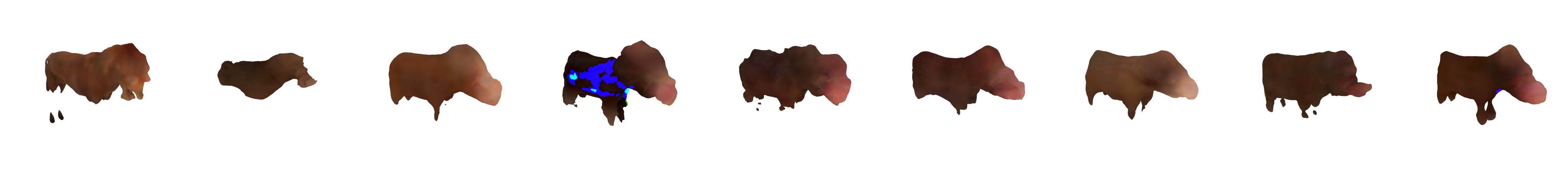}

\begin{minipage}[t] {\textwidth}
\vspace{-5mm}
\begin{adjustbox}{width=\textwidth}
    \begin{tabular}{>{\centering\arraybackslash} p{1cm}
    >{\centering\arraybackslash} p{1cm}
    >{\centering\arraybackslash} p{1cm}
    >{\centering\arraybackslash} p{1cm}
    >{\centering\arraybackslash} p{1cm}
    >{\centering\arraybackslash} p{1cm}
    >{\centering\arraybackslash} p{1cm}
    >{\centering\arraybackslash} p{1cm}
    >{\centering\arraybackslash} p{1cm}
    >{\centering\arraybackslash} p{1cm}
    >{\centering\arraybackslash} p{1cm}}
          \tiny{Input } &   \tiny{No Aug.} &   \tiny{Random Aff.} &   \tiny{N. Cage~\cite{Yifan:NeuralCage:2020}} &   \tiny{Random COCO~\cite{coco}} &   \tiny{S. Trans.~\cite{Li_2019_CVPR}} & \tiny{N. C.~\cite{Yifan:NeuralCage:2020} \& S. T.~\cite{Li_2019_CVPR}}  &  \tiny{Our Geo} &   \tiny{Our Tex} & \tiny{Our Geo + Tex}
    \end{tabular}
\end{adjustbox}
\end{minipage}
\vspace{-5mm}
\caption{\small Qualitative results on {\bf Single Image 3D reconstruction} using DISN as the 3D reconstruction method and various 3D data augmentation strategies. While none of the results are perfect, some are clearly worse than others. Affine randomization hurts performance. No augmentation produces worst results than the remaining augmentation strategies. Ours produces the most plausible and smooth shapes.	\vspace{-6mm}}
\label{fig:single-image-3d}
\vspace{-0mm}
\end{figure*}

\vspace{-2mm}
\subsection{3D Data Augmentation via {\ourmodel}}\label{ssec:exp:dataaug}
\vspace{-2mm}
Training neural networks to reconstruct 3D objects from partial observations such as from a single monocular image, requires large amounts of training data. However, obtaining very large-scale 3D object datasets with high quality geometry and texture is  hard and expensive. To augment the 3D training set, the most widely used technique is domain randomization~\cite{DR17} which randomizes colors/textures of objects when training downstream models. 
In our work, we propose to use our method as a way to perform 3D synthetic data augmentation. In particular, we here focus on the task of single-image 3D reconstruction of single objects.  
We propose to use both geometry and texture style transfer to augment the 3D training data, and validate the performance of our model with comparisons to strong baselines. 

\textbf{Experimental Settings:}
As described in \S\ref{ssec:exp:dataset}, we use 221 animal objects randomly selected from the 442 collected animals from Turbosquid as our training set for single image 3D reconstruction. We use our {\ourmodel} to augment the training set by transferring  the geometry and texture from one 3D object to another, which yields $221\times 221$  objects in total. We render all objects into 24 different views, and train DISN~\cite{disn} to predict 3D shapes from images. We additionally predict RGB color for each 3D coordinate. Training details can be found in the Supplement.
To evaluate performance, we use the remaining 221 objects from Turbosquid as test set, and split them into three categories: Seen Shapes, Similar Shapes, and Unseen Shapes according to the chamfer distance to the closest shape in the training set(before augmentation).  Note that while the shape of the objects in Seen category matches some in the training set, the texture of the objects is distinct. Following past literature~\cite{occnet, disn, gao2020deftet,Kaolin19}, we evaluate the 3D reconstruction quality using the Chamfer Distance, Chamfer-L1, and F-score. We provide  results in terms of other metrics and results on another category(Cars) in Supplement. 

To evaluate our method as a data augmentation strategy, we compare with the following baselines: {\bf a)} no data augmentation, where the network is only trained on (rendered views of) 221 training objects. For texture-based augmentation, we compare with {\bf b)} random texture replacement akin to domain randomization work~\cite{DR17}, where for each of the 221 objects we randomly select $221$ COCO~\cite{coco} images as new textures (yielding $221\times 221$ combinations), and {\bf c)} using image style transfer~\cite{Li_2019_CVPR} to transfer the texture image style from one training object to another. For geometry-based augmentation, we compare with {\bf d)} random affine transformations, where we randomly apply different $221$ affine transformations to each of the training objects (scale of the affine transformation is the same as in our method), and {\bf e)} using a Neural Cage model~\cite{Yifan:NeuralCage:2020} trained on our 1,120 non-textured shape set to deform the geometry of each training object to another. We further compare with {\bf f)} a combination of Neural Cage for shape deformation and Image Style Transfer~\cite{Li_2019_CVPR}  for texture augmentation. Finally, we also ablate our own method when using geometry style transfer only,  texture style transfer only, and both texture and geometry but without fine-tuning stage. Note that across all baselines but the no-data-augmentation, we use $221\times 221$ textured shapes for training the 3D reconstruction method, making this a fair experiment.

\begin{figure}[t!]
\vspace{-0mm}
\centering
\includegraphics[width=0.2\textwidth]{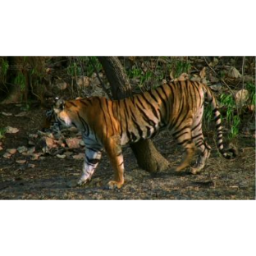}
\includegraphics[width=0.28\textwidth,trim=0 50 1536 100,clip]{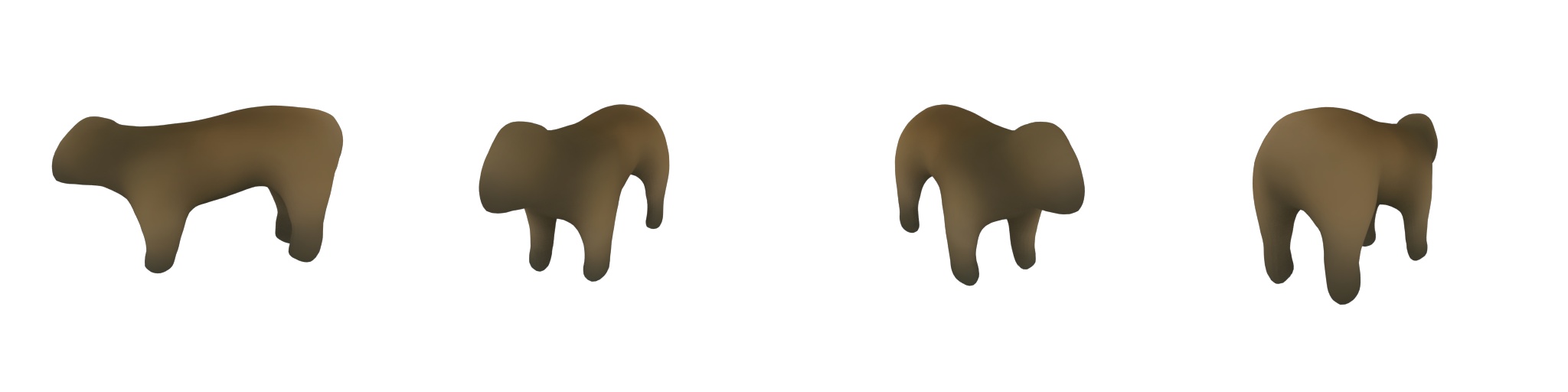}
\includegraphics[width=0.2\textwidth]{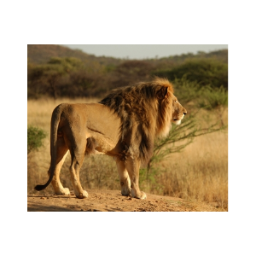}
\includegraphics[width=0.28\textwidth,trim=0 20 1536 100,clip]{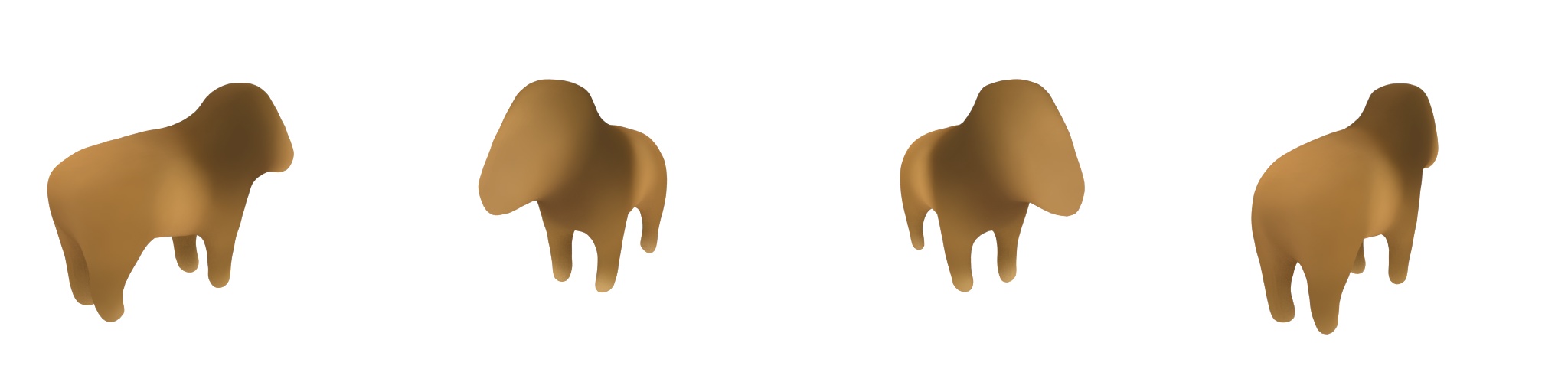}
\includegraphics[width=0.2\textwidth]{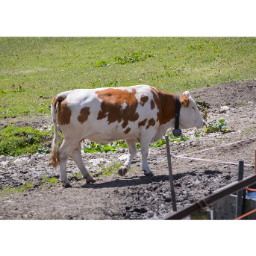}
\includegraphics[width=0.28\textwidth,trim=0 100 1536 50,clip]{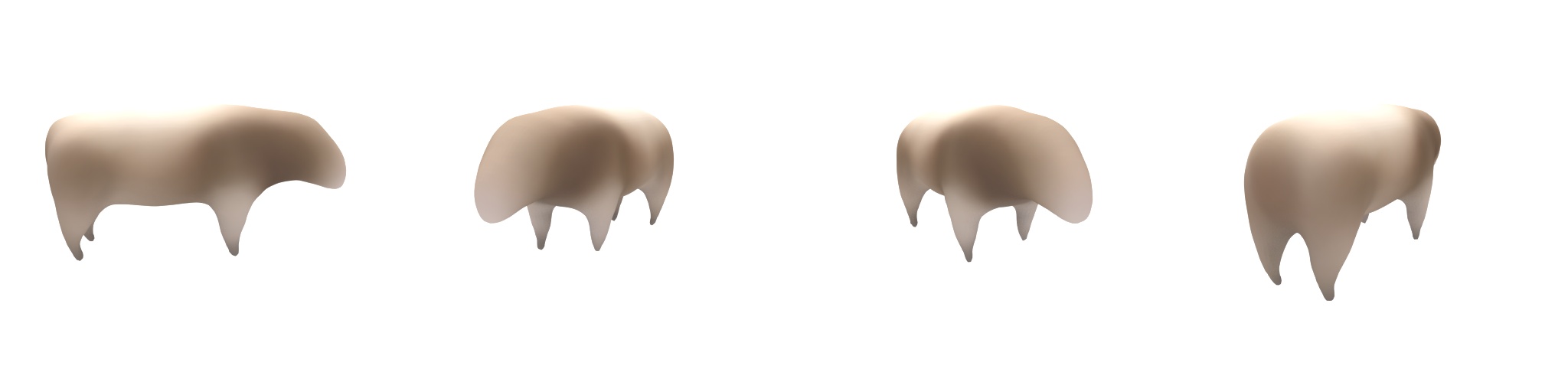}
\includegraphics[width=0.2\textwidth]{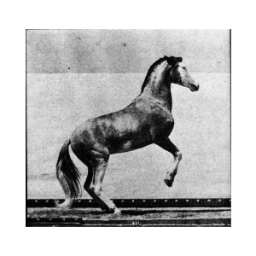}
\includegraphics[width=0.28\textwidth,trim=0 100 1536 10,clip]{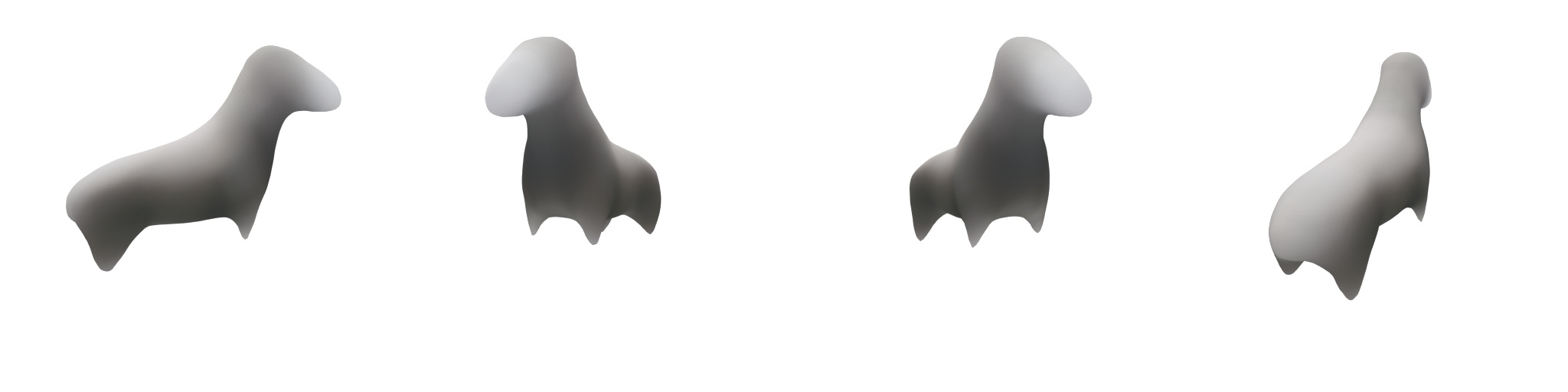}

\vspace{-4mm}
\caption{\small Qualitative results on {\bf Single Image 3D reconstruction} on SMAL~\cite{SMAL} Dataset using DISN~\cite{disn} as the 3D reconstruction method and {\ourmodel} as a 3D data augmentation strategy. 	Note that the background is masked out with the provided segmentation in the Dataset.	\vspace{-6mm}}
\label{fig:single-image-3d-real}
\end{figure}

\textbf{Experimental Results:}
Quantitative and qualitative results are shown in Table~\ref{tbl:single-image-3d-chamfer} and Fig.~\ref{fig:single-image-3d}, respectively. 
Comparing with no data augmentation which is typically done in 3D reconstruction works, ours and baseline augmentation methods achieve significant improvements.
Random Affine transformations achieves the worst results, as it degrades the quality of training data and makes it harder for the networks to learn shape priors. Comparing the Neural Cage~\cite{Yifan:NeuralCage:2020} Geometry with Ours Geometry Transfer only, we achieve higher quality 3D reconstruction, showing the usefulness of affine transformation field in geometric style transfer. For texture augmentation, compared to both Random COCO~\cite{coco} texture and Style Transfer~\cite{Li_2019_CVPR} Texture, ours texture transfer method achieves comparable or in some metrics worst performance. Note that the texture-based augmentation methods perform best on the Seen category, which we attribute to the network overfitting to the geometry in the training set. Since some test shapes in seen category have the same shape geometry but different texture as in training set, the texture randomization methods are directly optimizing for the downstream network to disregard texture and memorize shapes.  The performance of texture-randomization methods drops significantly on the Unseen category where both textures and shapes are novel. Here our full approach significantly outperforms texture-randomization methods. Comparing Neural Cage~\cite{Yifan:NeuralCage:2020} + Style Transfer~\cite{Li_2019_CVPR} with our full model, we achieve comparable performance in terms of Chamfer and Chamfer-L1, and better performance in terms of F-score and qualitative results. This demonstrates the effectiveness of {\ourmodel} as a 3D data augmentation technique.

\textbf{Qualitative Results on Real Data:} 
To evaluate how well the 3D reconstruction model trained using our augmentation strategy generalizes to real images, we directly utilize the trained network to the SMAL~\cite{SMAL} dataset without fine-tuning. Since DISN~\cite{disn} requires camera pose during inference, we train the occupancy network that does not require the camera views on our augmented $221\times 221$ data. We provide qualitative results in Fig.~\ref{fig:single-image-3d-real}. Although we only train the model on synthetic rendered images and human-created shapes from Turbosquid, the network is able to reconstruct shapes from masked real images to a good precision. 

Additional results can be found in Supplement.

\vspace{-3mm}
\section{Conclusion}
\vspace{-2mm}
In this paper, we proposed a novel method for 3D object stylization informed by a reference textured 3D shape. Our {\ourmodel} predicts a part-aware affine transformation field that warps a source shape to imitate the overall geometric style of the target shape. We also transfer the texture style of the target  to the source object with the help of a multi-view differentiable renderer and the  geometric alignment after shape stylization. 
Our method jointly optimizes the geometric style network and an image style transfer network with losses defined over both the geometry and the multi-view rendering of a pair of textured shapes.   
We showcase our approach   on 3D content stylization,  as well as a valuable tool to create 3D data augmentations for computer vision tasks. We outperform traditional augmentation techniques, particularly on the challenging shapes unseen at training time. We hope that our work opens an avenue to creative 3D content stylization and creation tooling for both naive and expert users.


{\small
\bibliographystyle{ieee_fullname}
\bibliography{reference}

\begin{thebibliography}{10}\itemsep=-1pt

\bibitem{wikiart}
Painter by numbers, wikiart.
\newblock \url{https://www.kaggle.com/c/painter-by-numbers}.

\bibitem{agarwala2004keyframe}
Aseem Agarwala, Aaron Hertzmann, David~H Salesin, and Steven~M Seitz.
\newblock Keyframe-based tracking for rotoscoping and animation.
\newblock {\em ACM Transactions on Graphics (ToG)}, 23(3):584--591, 2004.

\bibitem{chang2015shapenet}
Angel~X Chang, Thomas Funkhouser, Leonidas Guibas, Pat Hanrahan, Qixing Huang,
  Zimo Li, Silvio Savarese, Manolis Savva, Shuran Song, Hao Su, et~al.
\newblock Shapenet: An information-rich 3d model repository.
\newblock {\em arXiv preprint arXiv:1512.03012}, 2015.

\bibitem{chen2019bae_net}
Zhiqin Chen, Kangxue Yin, Matthew Fisher, Siddhartha Chaudhuri, and Hao Zhang.
\newblock Bae-net: Branched autoencoder for shape co-segmentation.
\newblock {\em Proceedings of International Conference on Computer Vision
  (ICCV)}, 2019.

\bibitem{deng2009imagenet}
Jia Deng, Wei Dong, Richard Socher, Li-Jia Li, Kai Li, and Li Fei-Fei.
\newblock Imagenet: A large-scale hierarchical image database.
\newblock In {\em 2009 IEEE conference on computer vision and pattern
  recognition}, pages 248--255. Ieee, 2009.

\bibitem{edelstein2020enigma}
Michal Edelstein, Danielle Ezuz, and Mirela Ben-Chen.
\newblock Enigma: evolutionary non-isometric geometry matching.
\newblock {\em ACM Transactions on Graphics (TOG)}, 39(4), 2020.

\bibitem{ezuz2019elastic}
Danielle Ezuz, Behrend Heeren, Omri Azencot, Martin Rumpf, and Mirela Ben-Chen.
\newblock Elastic correspondence between triangle meshes.
\newblock In {\em Computer Graphics Forum}, volume~38, pages 121--134. Wiley
  Online Library, 2019.

\bibitem{gal2009iwires}
Ran Gal, Olga Sorkine, Niloy~J Mitra, and Daniel Cohen-Or.
\newblock iwires: An analyze-and-edit approach to shape manipulation.
\newblock In {\em ACM SIGGRAPH 2009 papers}, pages 1--10. 2009.

\bibitem{gal20073d}
Ran Gal, Olga Sorkine, Tiberiu Popa, Alla Sheffer, and Daniel Cohen-Or.
\newblock 3d collage: expressive non-realistic modeling.
\newblock In {\em Proceedings of the 5th international symposium on
  Non-photorealistic animation and rendering}, pages 7--14, 2007.

\bibitem{gao2020deftet}
Jun Gao, Wenzheng Chen, Tommy Xiang, Clement~Fuji Tsang, Alec Jacobson, Morgan
  McGuire, and Sanja Fidler.
\newblock Learning deformable tetrahedral meshes for 3d reconstruction.
\newblock In {\em Advances In Neural Information Processing Systems}, 2020.

\bibitem{gao2018automatic}
Lin Gao, Jie Yang, Yi-Ling Qiao, Yu-Kun Lai, Paul~L Rosin, Weiwei Xu, and
  Shihong Xia.
\newblock Automatic unpaired shape deformation transfer.
\newblock {\em ACM Transactions on Graphics (TOG)}, 37(6):1--15, 2018.

\bibitem{gatys2016image}
Leon~A Gatys, Alexander~S Ecker, and Matthias Bethge.
\newblock Image style transfer using convolutional neural networks.
\newblock In {\em Proceedings of the IEEE Conference on Computer Vision and
  Pattern Recognition}, pages 2414--2423, 2016.

\bibitem{Gatys2017ControllingPF}
Leon~A. Gatys, Alexander~S. Ecker, Matthias Bethge, Aaron Hertzmann, and Eli
  Shechtman.
\newblock Controlling perceptual factors in neural style transfer.
\newblock {\em 2017 IEEE Conference on Computer Vision and Pattern Recognition
  (CVPR)}, pages 3730--3738, 2017.

\bibitem{gooch2001non}
Bruce Gooch and Amy Gooch.
\newblock {\em Non-photorealistic rendering}.
\newblock CRC Press, 2001.

\bibitem{hertz2020deep}
Amir Hertz, Rana Hanocka, Raja Giryes, and Daniel Cohen-Or.
\newblock Deep geometric texture synthesis.
\newblock {\em arXiv preprint arXiv:2007.00074}, 2020.

\bibitem{hertzmann1998painterly}
Aaron Hertzmann.
\newblock Painterly rendering with curved brush strokes of multiple sizes.
\newblock In {\em Proceedings of the 25th annual conference on Computer
  graphics and interactive techniques}, pages 453--460, 1998.

\bibitem{Hertzmann2001ImageA}
Aaron Hertzmann, Charles~E. Jacobs, Nuria Oliver, Brian Curless, and David
  Salesin.
\newblock Image analogies.
\newblock In {\em SIGGRAPH}, 2001.

\bibitem{hoppe1996progressive}
Hugues Hoppe.
\newblock Progressive meshes.
\newblock In {\em Proceedings of the 23rd annual conference on Computer
  graphics and interactive techniques}, pages 99--108, 1996.

\bibitem{Hu2017Style}
Ruizhen Hu, Wenchao Li, Oliver~Van Kaick, Hui Huang, Melinos Averkiou, Daniel
  Cohen-Or, and Hao Zhang.
\newblock Co-locating style-defining elements on 3d shapes.
\newblock {\em ACM Transactions on Graphics}, 36(3):33:1--33:15, June 2017.

\bibitem{Huang2017ArbitraryST}
Xun Huang and Serge~J. Belongie.
\newblock Arbitrary style transfer in real-time with adaptive instance
  normalization.
\newblock {\em 2017 IEEE International Conference on Computer Vision (ICCV)},
  pages 1510--1519, 2017.

\bibitem{Kaolin19}
Krishna~Murthy Jatavallabhula, Edward Smith, Jean-Francois Lafleche,
  Clement~Fuji Tsang, Artem Rozantsev, Wenzheng Chen, Tommy Xiang, Rev
  Lebaredian, and Sanja Fidler.
\newblock Kaolin: A pytorch library for accelerating 3d deep learning research.
\newblock In {\em arXiv:1911.05063}, 2019.

\bibitem{jing2017neural}
Yongcheng Jing, Yezhou Yang, Zunlei Feng, Jingwen Ye, Yizhou Yu, and Mingli
  Song.
\newblock Neural style transfer: A review.
\newblock {\em arXiv preprint arXiv:1705.04058}, 2017.

\bibitem{stewart17}
Bhautik Joshi, Kristen Stewart, and David Shapiro.
\newblock Bringing impressionism to life with neural style transfer in come
  swim.
\newblock {\em arXiv preprint arXiv:1701.04928}, 2017.

\bibitem{kato2018renderer}
Hiroharu Kato, Yoshitaka Ushiku, and Tatsuya Harada.
\newblock Neural 3d mesh renderer.
\newblock In {\em The IEEE Conference on Computer Vision and Pattern
  Recognition (CVPR)}, 2018.

\bibitem{Kyprianidis2013StateOT}
Jan~Eric Kyprianidis, John~P. Collomosse, Tinghuai Wang, and Tobias Isenberg.
\newblock State of the "art": A taxonomy of artistic stylization techniques for
  images and video.
\newblock {\em IEEE transactions on visualization and computer graphics}, 19
  5:866--85, 2013.

\bibitem{Laine2020diffrast}
Samuli Laine, Janne Hellsten, Tero Karras, Yeongho Seol, Jaakko Lehtinen, and
  Timo Aila.
\newblock Modular primitives for high-performance differentiable rendering.
\newblock {\em ACM Transactions on Graphics}, 39(6), 2020.

\bibitem{Li_2019_CVPR}
Xueting Li, Sifei Liu, Jan Kautz, and Ming-Hsuan Yang.
\newblock Learning linear transformations for fast image and video style
  transfer.
\newblock In {\em The IEEE Conference on Computer Vision and Pattern
  Recognition (CVPR)}, June 2019.

\bibitem{Li2017UniversalST}
Yijun Li, Chen Fang, Jimei Yang, Zhaowen Wang, Xin Lu, and Ming-Hsuan Yang.
\newblock Universal style transfer via feature transforms.
\newblock In {\em NIPS}, 2017.

\bibitem{Li2018ACS}
Yijun Li, Ming-Yu Liu, Xueting Li, Ming-Hsuan Yang, and Jan Kautz.
\newblock A closed-form solution to photorealistic image stylization.
\newblock {\em CoRR}, abs/1802.06474, 2018.

\bibitem{coco}
Tsung-Yi Lin, Michael Maire, Serge Belongie, James Hays, Pietro Perona, Deva
  Ramanan, Piotr Doll{\'a}r, and C~Lawrence Zitnick.
\newblock Microsoft coco: Common objects in context.
\newblock In {\em European conference on computer vision}, pages 740--755.
  Springer, 2014.

\bibitem{lindholm2001user}
Erik Lindholm, Mark~J Kilgard, and Henry Moreton.
\newblock A user-programmable vertex engine.
\newblock In {\em Proceedings of the 28th annual conference on Computer
  graphics and interactive techniques}, pages 149--158. ACM, 2001.

\bibitem{liu_2019}
Hsueh-Ti~Derek Liu and Alec Jacobson.
\newblock Cubic stylization.
\newblock {\em ACM Transactions on Graphics (TOG)}, 38(6):1–10, Nov 2019.

\bibitem{liu2020neural}
Hsueh-Ti~Derek Liu, Vladimir~G Kim, Siddhartha Chaudhuri, Noam Aigerman, and
  Alec Jacobson.
\newblock Neural subdivision.
\newblock {\em ACM Transactions on Graphics (TOG)}, 39(4):124--1, 2020.

\bibitem{liu2018paparazzi}
Hsueh-Ti~Derek Liu, Michael Tao, and Alec Jacobson.
\newblock Paparazzi: surface editing by way of multi-view image processing.
\newblock {\em ACM Trans. Graph.}, 37(6):221--1, 2018.

\bibitem{liu2019pvcnn}
Zhijian Liu, Haotian Tang, Yujun Lin, and Song Han.
\newblock Point-voxel cnn for efficient 3d deep learning.
\newblock In {\em Advances in Neural Information Processing Systems}, 2019.

\bibitem{luo2015legolization}
Sheng-Jie Luo, Yonghao Yue, Chun-Kai Huang, Yu-Huan Chung, Sei Imai, Tomoyuki
  Nishita, and Bing-Yu Chen.
\newblock Legolization: Optimizing lego designs.
\newblock {\em ACM Transactions on Graphics (TOG)}, 34(6):1--12, 2015.

\bibitem{occnet}
Lars Mescheder, Michael Oechsle, Michael Niemeyer, Sebastian Nowozin, and
  Andreas Geiger.
\newblock Occupancy networks: Learning 3d reconstruction in function space.
\newblock In {\em Proceedings IEEE Conf. on Computer Vision and Pattern
  Recognition (CVPR)}, 2019.

\bibitem{mordvintsev2018differentiable}
Alexander Mordvintsev, Nicola Pezzotti, Ludwig Schubert, and Chris Olah.
\newblock Differentiable image parameterizations.
\newblock {\em Distill}, 2018.
\newblock https://distill.pub/2018/differentiable-parameterizations.

\bibitem{panozzo2013weighted}
Daniele Panozzo, Ilya Baran, Olga Diamanti, and Olga Sorkine-Hornung.
\newblock Weighted averages on surfaces.
\newblock {\em ACM Transactions on Graphics (TOG)}, 32(4):1--12, 2013.

\bibitem{schmidt2019distortion}
Patrick Schmidt, Janis Born, Marcel Campen, and Leif Kobbelt.
\newblock Distortion-minimizing injective maps between surfaces.
\newblock {\em ACM Transactions on Graphics (TOG)}, 38(6):1--15, 2019.

\bibitem{schmidt2020inter}
Patrick Schmidt, Marcel Campen, Janis Born, and Leif Kobbelt.
\newblock Inter-surface maps via constant-curvature metrics.
\newblock {\em ACM Transactions on Graphics (TOG)}, 39(4):119--1, 2020.

\bibitem{shen2012sd}
Liang-Tsen Shen, Sheng-Jie Luo, Chun-Kai Huang, and Bing-Yu Chen.
\newblock Sd models: Super-deformed character models.
\newblock In {\em Computer Graphics Forum}, volume~31, pages 2067--2075. Wiley
  Online Library, 2012.

\bibitem{Sheng2018AvatarNetMZ}
Lu Sheng, Ziyi Lin, Jing Shao, and Xiaogang Wang.
\newblock Avatar-net: Multi-scale zero-shot style transfer by feature
  decoration.
\newblock In {\em Proceedings of the IEEE Conference on Computer Vision and
  Pattern Recognition}, pages 8242--8250, 2018.

\bibitem{simonyan2014very}
Karen Simonyan and Andrew Zisserman.
\newblock Very deep convolutional networks for large-scale image recognition.
\newblock {\em arXiv preprint arXiv:1409.1556}, 2014.

\bibitem{sorkine2005laplacian}
Olga Sorkine.
\newblock Laplacian mesh processing.
\newblock {\em Eurographics (STARs)}, 29, 2005.

\bibitem{sorkine2007rigid}
Olga Sorkine and Marc Alexa.
\newblock As-rigid-as-possible surface modeling.
\newblock In {\em Symposium on Geometry processing}, volume~4, pages 109--116,
  2007.

\bibitem{stein2018developability}
Oded Stein, Eitan Grinspun, and Keenan Crane.
\newblock Developability of triangle meshes.
\newblock {\em ACM Transactions on Graphics (TOG)}, 37(4):1--14, 2018.

\bibitem{sumner2004deformation}
Robert~W Sumner and Jovan Popovi{\'c}.
\newblock Deformation transfer for triangle meshes.
\newblock {\em ACM Transactions on graphics (TOG)}, 23(3):399--405, 2004.

\bibitem{sumner2007embedded}
Robert~W Sumner, Johannes Schmid, and Mark Pauly.
\newblock Embedded deformation for shape manipulation.
\newblock In {\em ACM SIGGRAPH 2007 papers}, pages 80--es. 2007.

\bibitem{takikawa2021neural}
Towaki Takikawa, Joey Litalien, Kangxue Yin, Karsten Kreis, Charles Loop, Derek
  Nowrouzezahrai, Alec Jacobson, Morgan McGuire, and Sanja Fidler.
\newblock Neural geometric level of detail: Real-time rendering with implicit
  3d shapes.
\newblock {\em arXiv preprint arXiv:2101.10994}, 2021.

\bibitem{theobalt2007animation}
Christian Theobalt, Christian R{\"o}ssl, Edilson de Aguiar, and Hans-Peter
  Seidel.
\newblock Animation collage.
\newblock In {\em Proceedings of the 2007 ACM SIGGRAPH/Eurographics symposium
  on Computer animation}, pages 271--280. Citeseer, 2007.

\bibitem{DR17}
J. {Tobin}, R. {Fong}, A. {Ray}, J. {Schneider}, W. {Zaremba}, and P. {Abbeel}.
\newblock Domain randomization for transferring deep neural networks from
  simulation to the real world.
\newblock In {\em IEEE/RSJ International Conference on Intelligent Robots and
  Systems (IROS)}, pages 23--30, 2017.

\bibitem{wang2016unsupervised}
Tuanfeng~Y Wang, Hao Su, Qixing Huang, Jingwei Huang, Leonidas~J Guibas, and
  Niloy~J Mitra.
\newblock Unsupervised texture transfer from images to model collections.
\newblock {\em ACM Trans. Graph.}, 35(6):177--1, 2016.

\bibitem{xu_siga10}
Kai Xu, Honghua Li, Hao Zhang, Daniel Cohen-Or, Yueshan Xiong, and Zhiquan
  Cheng.
\newblock Style-content separation by anisotropic part scales.
\newblock {\em ACM Transactions on Graphics (Proc. of SIGGRAPH Asia 2010)},
  29(5):184:1--184:10, 2010.

\bibitem{disn}
Qiangeng Xu, Weiyue Wang, Duygu Ceylan, Radomir Mech, and Ulrich Neumann.
\newblock Disn: Deep implicit surface network for high-quality single-view 3d
  reconstruction.
\newblock In {\em Advances in Neural Information Processing Systems 32}, pages
  492--502. Curran Associates, Inc., 2019.

\bibitem{yi2017large}
Li Yi, Lin Shao, Manolis Savva, Haibin Huang, Yang Zhou, Qirui Wang, Benjamin
  Graham, Martin Engelcke, Roman Klokov, Victor Lempitsky, et~al.
\newblock Large-scale 3d shape reconstruction and segmentation from shapenet
  core55.
\newblock {\em arXiv preprint arXiv:1710.06104}, 2017.

\bibitem{Yifan:NeuralCage:2020}
Wang Yifan, Noam Aigerman, Vladimir~G. Kim, Siddhartha Chaudhuri, and Olga
  Sorkine-Hornung.
\newblock Neural cages for detail-preserving 3d deformations.
\newblock In {\em CVPR}, 2020.

\bibitem{SMAL}
Silvia Zuffi, Angjoo Kanazawa, David Jacobs, and Michael~J. Black.
\newblock {3D} menagerie: Modeling the {3D} shape and pose of animals.
\newblock In {\em IEEE Conf. on Computer Vision and Pattern Recognition
  (CVPR)}, July 2017.

\end{thebibliography}
}

\end{document}